\documentclass[preprint,12pt, a4paper]{elsarticle}

\usepackage{amssymb}
\usepackage{hyperref}
\usepackage{siunitx}
\usepackage{graphicx}
\usepackage{listings}
\usepackage{soul}
\usepackage{todonotes}
\usepackage{multirow}
\usepackage{adjustbox}
\usepackage{booktabs}

\newcommand{\pop}[1]{}

\lstdefinestyle{prompt}{
    basicstyle=\ttfamily,
columns=flexible,
breaklines=true,
frame=single
}

\DeclareFixedFont{\ttb}{T1}{txtt}{bx}{n}{12} 
\DeclareFixedFont{\ttm}{T1}{txtt}{m}{n}{12}  

\usepackage{color}
\definecolor{deepblue}{rgb}{0,0,0.5}
\definecolor{deepred}{rgb}{0.6,0,0}
\definecolor{deepgreen}{rgb}{0,0.5,0}

\newcommand\pythonstyle{\lstset{
language=Python,
basicstyle=\footnotesize\ttfamily,
morekeywords={self},              
keywordstyle=\ttb\color{deepblue},
emph={MyClass,__init__},          
emphstyle=\ttb\color{deepred},    
stringstyle=\color{deepgreen},
frame=tb,                         
showstringspaces=false,
breaklines=true
}}

\lstnewenvironment{python}[1][]
{
\pythonstyle
\lstset{#1}
}
{}

\journal{SoftwareX}

\begin{document}
\renewcommand{\labelenumii}{\arabic{enumi}.\arabic{enumii}}


\begin{frontmatter}
 


\title{llmNER: (Zero$|$Few)-Shot Named Entity Recognition, Exploiting the Power of Large Language Models}


\author[dccuchile,imfd]{Fabián Villena}
\ead{fvillena@imfd.cl}
\author[imfd,dccuc]{Luis Miranda}
\ead{lmirandn@uc.cl}
\author[imfd,fcfm]{Claudio Aracena}
\ead{claudio.aracena@uchile.cl}

\address[dccuchile]{Department of Computer Science, University of Chile, Chile}
\address[imfd]{Millennium Institute for Foundational Research on Data, Chile}
\address[dccuc]{Department of Computer Science, Catholic University of Chile, Chile}
\address[fcfm]{Faculty of Physical and Mathematical Sciences, University of Chile, Chile}

\begin{abstract}
Large language models (LLMs) allow us to generate high-quality human-like text. One interesting task in natural language processing (NLP) is named entity recognition (NER), which seeks to detect mentions of relevant information in documents. This paper presents \texttt{llmNER}, a Python library for implementing zero-shot and few-shot NER with LLMs; by providing an easy-to-use interface, \texttt{llmNER} can compose prompts, query the model, and parse the completion returned by the LLM. Also, the library enables the user to perform prompt engineering efficiently by providing a simple interface to test multiple variables. We validated our software on two NER tasks to show the library's flexibility. \texttt{llmNER} aims to push the boundaries of in-context learning research by removing the barrier of the prompting and parsing steps.
\end{abstract}


\end{frontmatter}


\section*{Metadata}
\label{}

\begin{table}[!h]
\begin{tabular}{|l|p{6.5cm}|p{6.5cm}|}
\hline
\textbf{Nr.} & \textbf{Code metadata description} & \textbf{} \\
\hline
C1 & Current code version & 0.1.0 \\
\hline
C2 & Permanent link to code/repository used for this code version & \url{https://github.com/plncmm/llmner} \\
\hline
C3  & Permanent link to Reproducible Capsule & \url{https://github.com/plncmm/llmner/blob/main/notebooks/1-example.ipynb}\\
\hline
C4 & Legal Code License   & Apache License, Version 2.0 \\
\hline
C5 & Code versioning system used & git \\
\hline
C6 & Software code languages, tools, and services used & python, LangChain \\
\hline
C7 & Compilation requirements, operating environments \& dependencies & Python $>=$ 3.10, git\\
\hline
C8 & If available Link to developer documentation/manual & \url{https://github.com/plncmm/llmner/blob/main/README.md/} \\
\hline
C9 & Support email for questions & fvillena@proton.me \\
\hline
\end{tabular}
\caption{Code metadata (mandatory)}
\label{codeMetadata} 
\end{table}

\section{Motivation and significance}
NLP has gained tremendous importance in recent years with the advent of Transformer-based pre-trained language models (PLM) \cite{Qiu2020}. These language models (LM) have become the new paradigm for NLP-based machine learning modeling because of their modularity and ease of transferring learning. Nowadays, one can fine-tune a off-the-shelf PLM to solve any NLP task using off-the-shelf \cite{dodge2020finetuning}. 

PLMs trained on web-scale unannotated text, such as BERT \cite{devlin-etal-2019-bert} and BERT-alike models, such as RoBERTa \cite{he2021deberta} and DeBERTa \cite{he2021deberta} have been the \textit{de facto} standard for solving NLP tasks. These pre-trained Transformer-based context-aware models have shown outstanding performance, but their size limits capacity. LLMs are PLMs with a significantly larger model size scale \cite{zhao2023survey}; for example, BERT has a model size of \num{0.3e9} parameters, and the LLM GPT-3 \cite{brown2020}, has \num{175e9} parameters. Additionally, it has been found that scaling PLMs improves the performance of the models on downstream tasks \cite{kaplan2020}.

Besides superior performance in downstream tasks, LLMs show surprising and more important behaviors in solving complex tasks, called emergent abilities. Emergent abilities are aptitudes not present in small models but arise in LLMs \cite{wei2022emergent} and include in-context learning, where a model can generate expected outputs to natural language instructions without additional training; instruction following, where a model fine-tuned using natural language instructions performs well on unseen tasks that are also described in the form of instructions; and \textit{step-by-step reasoning}, where a model can solve complex problems by instructing the model involving intermediate reasoning steps for deriving the final answer.

In-context learning (ICL) is a paradigm that allows language models to learn tasks given only a few examples in the form of demonstration. Although LLMs themselves show good ICL capabilities, some techniques, such as supervised instruction-tuning, are used to improve ICL performance. Instruction tuning enhances ICL model capabilities by updating model parameters by fine-tuning the model on task instructions \cite{dong2023survey}.

Few-shot learning is a machine learning problem where the model only has limited examples with supervised information to solve the task \cite{wang2020generalizing}. This branch of task-agnostic pre-training and task-agnostic model architectures has recently proliferated in the NLP area. By exploiting ICL, an LM can be contextualized to solve a task by only giving zero, one, or few examples without fine-tuning. This paradigm of solving a task with few to no examples solves the issue of the need for large datasets for training models for new tasks but with a significant penalty in performance compared to state-of-the-art fine-tuned models \cite{brown2020}.

NER is the task of finding spans of text that constitute named entities (anything that can be referred to with a proper name) and tagging the entity type. The four most common entity types are person, location, organization, and geopolitical entity. Sometimes, the term named entity is extended to include things that are not entities, including dates, times, and even numerical expressions \cite{jurafsky}.

For training NER models, one needs a large \textit{corpus} of task-specific annotated text, but constructing an annotated \textit{corpus} is both time-consuming and expensive. Also, after the \textit{corpus} is annotated, one needs to fine-tune a model (only if a foundation model is available for the domain), a process that is also expensive. We propose exploiting LLMs' ICL ability to solve NER tasks using zero and few-shot learning without \textit{corpora} annotation and model fine-tuning. Zero and few-shot learning NER could be helpful when no annotated \textit{corpora} are available; no base models are available to fine-tune; when one needs to prototype a NER-based system before spending time and money annotating \textit{corpora} and training models; or as a pre-annotation step to aid human annotation processes.

To leverage the ICL ability of LLMs, one must prompt the model using instructions in natural language to demonstrate the context of the task to the model. In the same way, the model will return a text completion also in natural language \cite{Liu2023}. The problem with this pipeline is that different prompts can lead to different results. Also, one must ensure a consistent output to parse the response into a machine-processable output. Ensuring a consistent output and the parsing of the completion can be wrapped into a high-level programming interface.

We propose \texttt{llmNER}, a Python library that wraps the prompting and completion parsing process into an easy-to-use interface. Our library implements multiple prompting methods, multiple answer shape parsers, a flexible prompt templating for adapting to other languages or domains, and the possibility to perform zero-shot and few-shot learning. This tool facilitates the process of prompt engineering for solving NER tasks.


There are many ways to prompt LLMs, and each method has a different impact on their performance and ability to solve specific tasks. For example, to perform machine translation, we must design the prompt to instruct the model to complete the input text with the explicit translation. In this case, we do not need to post-process the completion because the completion is the prediction itself. On the other hand, to perform text classification, we must design the prompt to make the model complete the text with a single word corresponding to an instance of the label space \cite{Liu2023}. Even though prompting methods for some NLP tasks, such as text classification \cite{yin-etal-2019-benchmarking,gao-etal-2021-making,hambardzumyan-etal-2021-warp,lester-etal-2021-power} or generation \cite{brown2020,Radford2019LanguageMA,schick-schutze-2021-shot}, prompting methods for NER still need to be explored. 

\citet{xie2023empirical} explored multiple prompting methods and only a JSON formatted answer shape for solving zero-shot NER tasks, where the most relevant are the following: a vanilla strategy, which we will call the \textbf{single-turn prompting method}, where they directly ask the model to annotate the entity mentions; and a decomposed-question-answering strategy, which we will call \textbf{multi-turn prompting method} where they broke down the task into simpler subproblems by asking the model to annotate the text one entity at a time and syntactic augmentation when they first ask the model to analyze the syntactic structure of the text and then perform NER. \citet{cui-etal-2021-template} used a different multi-turn prompting method, where the model scores candidate text spans based on how well they fit alongside the entity classes. Finally, \citet{wang2023gptner} used a single-turn prompting method, prompting the model to echo the input document with in-line annotations where special symbols enclosed the entity mentioned inside the input text.

NER remains a tough challenge for ICL, falling far behind \cite{pang-etal-2023-guideline, xie2023empirical, cui-etal-2021-template, wang2023gptner} state-of-the-art fine-tuned models \cite{wang-etal-2021-automated}, but shows capabilities for universal information extraction without training data \cite{dukić2024always}; therefore, a necessity arises to develop research to improve the results of this paradigm.

Multiple domain-specific applications of NER through ICL have been investigated; for example, in the biomedical and clinical domains \cite{Hu2024}, this paradigm has been used to extract information for cancer research \cite{Jiang2024}, rare diseases \cite{Shyr2024}, and adverse drug events \cite{Modi2024}; in the legal domain to extract contract \cite{Adibhatla2024} and legal violation \cite{bernsohn2024legallens} information; in the social media domain to detect offensive text \cite{Paraschiv2023} and in human resources domain to extract skill information \cite{nguyen2024rethinking}.

\section{Software description}

\texttt{llmNER} streamlines the prompting and answers parsing steps for NER through zero-shot and few-shot learning; in its most basic usage, the user only defines in natural language the entities to recognize, and then the library takes the model completion and parses the answer into a machine-readable abstract object. This tool also implements multiple prompting methods, answer shape parsers, and POS augmentation to streamline the prompt engineering process. The library's interface is based on the well-known machine learning library \texttt{scikit-learn} \cite{sklearn_api} to ease the learning process of using \texttt{llmNER}. As the backend LLM, the library can use any model platform that exposes an OpenAI-compatible API endpoint.

\subsection{Software architecture}
\begin{figure}[h]
\caption{Diagram of the architecture of the library and the pipeline for Named Entity Recognition.}
\centering
\vspace*{5mm}
\includegraphics[width=1\textwidth]{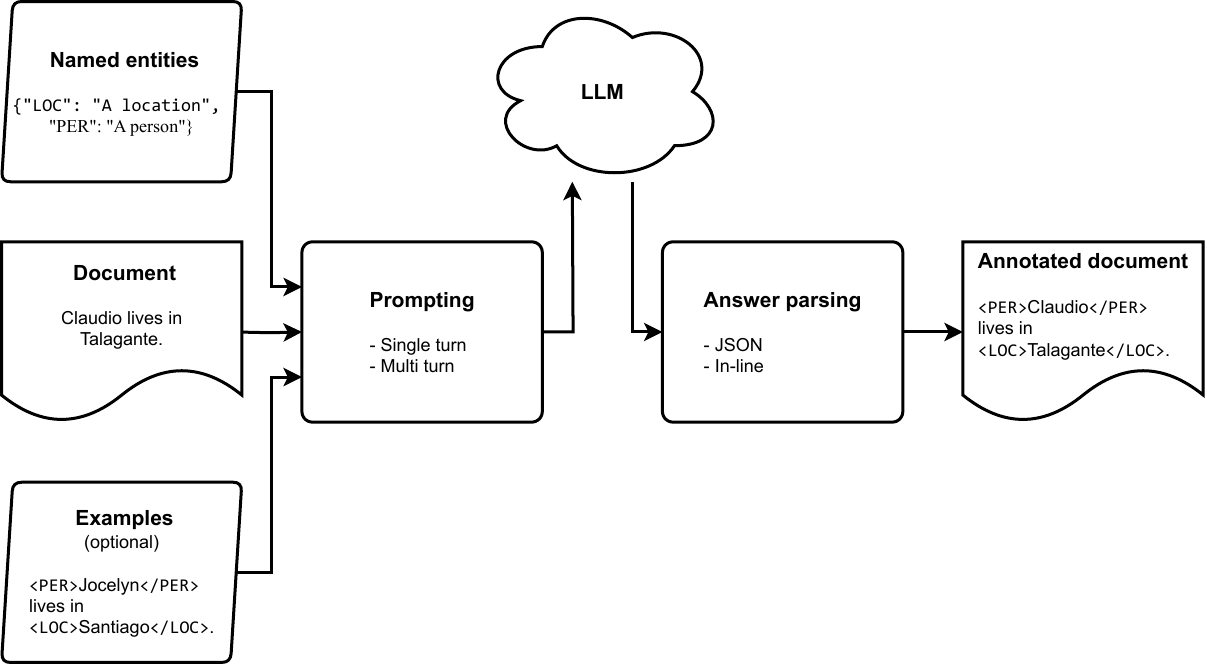}
\label{fig:pipeline}
\end{figure}

The library components are divided into the prompting methods and the answer parsers. The named entity definitions and few-shot examples provided by the user are compiled as a text prompt using a prompting method. Then the completion performed by the LLM is parsed using an answer parser, returning an object containing the named entity mentions. A diagram of this architecture is in Figure \ref{fig:pipeline}.

To assist the user in the prompt engineering phase of a zero-shot or few-shot learning NER pipeline, the library implements the following prompting methods:

\begin{description}
    \item[Single-turn] In this prompting method, the model is asked to annotate the entity mentions in a single response as in this summarized prompt: \textit{Annotate the mentions of the entities location and person: \texttt{\{response\}}}.
    \item[Multi-turn] In this prompting method, the model is asked to annotate one entity at a time \cite{xie2023empirical}, exploiting the chain-of-though following ability of LLMs, as in this summarized prompt: \textit{Annotate the mentions of the entity location: \texttt{\{response\}}. And now annotate the mentions of the entity person: \texttt{\{response\}}}.
    \begin{description}
        \item[Step-by-step annotation] In this option, the entity mentions are stored in a set at each turn, and at the end, this set is used as the final annotations. This option is useful when a specific text span can be associated with multiple entities. The user can also define the delimiters \cite{wang2023gptner} that enclose the entity mentions at each turn.
        \item[Final-step annotation] In this option, the model is prompted to annotate all the entities after the last turn, as in the single-turn prompting method. 
    \end{description}
\end{description}

Also, the library implements the following answer parsers:

\begin{description}
    \item[In-line] In this answer shape, the model is prompted to echo the exact input text, but with the entity mentions enclosed with in-line tags named after each entity class as in \texttt{<person>Fei-Fei Li</person> is a female scientist born in <country>China</country>}, where \textit{Fei-Fei Li} is an instance of the Person entity class, and \textit{China} is an instance of the Country entity class.
    \begin{description}
        \item[Custom delimiters] The in-line tags can be defined by the user. For example, the user can define \texttt{@@} as an opening tag and \texttt{\#\#} as a closing tag for an entity. This mode can be used only in a multi-turn setting. 
    \end{description}
    \item[JSON] In this answer shape, the model is prompted to compose the annotations as a JSON object, where each key is the entity name and each value is a list of text mentions, as in \texttt{\{"person":["Fei-Fei Li"], "country":["China"]\}}.
\end{description}

Finally, the user can augment the prompt with part-of-speech (POS) data. This syntactic information, by default, is added through ICL by prompting the model to add the data.

 \subsection{Software functionalities}

\texttt{llmNER} specialized in NER using the power of LLMs. Its main functionalities are:
\begin{itemize}
    \item Zero-shot learning for NER: the user can input a document (or multiple documents) and a list of entities to detect with their descriptions. \texttt{llmNER} will generate a prompt to query a selected LLM and parse the response to extract recognized entities.
    \item Few-shot learning for NER: Similar to the previous functionality, but the user can also input examples of NER to give a better context to the LLM when querying it. 
    \item There are multiple options for querying and parsing, such as single or multi-turn prompting and in-line or JSON parsers, as explained in the previous section.
    \item POS tags can augment the context for prompting a selected LLM. These tags can be generated by a function (from an ad-hoc library such as \texttt{NLTK} or \texttt{Spacy}) or by a prompt to a LLM.
    \item When the user passes multiple documents, the library can parallelize the querying process to get results faster. 
\end{itemize}
  
 \subsection{Sample code snippets analysis}

To display the ease of use of \texttt{llmNER}, Figure \ref{fig:example_usage1} and \ref{fig:example_usage2} show examples for zero-shot and few-shot cases, respectively.

\begin{figure}
	\begin{python}
# INPUT:

import os

os.environ["OPENAI_API_KEY"] = "<your OpenAI api key>"
from llmner import ZeroShotNer

entities = {
"person": "A person name, it can include first and last names, for example: John, Fabian or Mark Miranda",
    "organization": "An organization name, it can be a company, a government agency, etc.",
    "location": "A location name, it can be a city, a country, etc.",}

model = ZeroShotNer()
model.contextualize(entities=entities)

model.predict(["Fei-Fei Li is a female scientist born in China."])

# OUTPUT:

AnnotatedDocument(
        text="Fei-Fei Li is a female scientist born in China.",
        annotations={
            Annotation(start=0, end=10, label="person"),
            Annotation(start=41, end=46, label="location"),
        },
)

\end{python}
\caption{Example usage of \texttt{llmNER} for zero-shot NER.}
	\label{fig:example_usage1}
\end{figure}

\begin{figure}
	\begin{python}
# INPUT:

examples = [
    AnnotatedDocument(
        text="Elon Musk is the owner of the US company Tesla",
        annotations={
            Annotation(start=30, end=32, label="location"),
            Annotation(start=0, end=9, label="person"),
            Annotation(start=41, end=46, label="organization"),
        },
    ),
    AnnotatedDocument(
        text="Bill Gates is the owner of Microsoft",
        annotations={
            Annotation(start=0, end=10, label="person"),
            Annotation(start=27, end=36, label="organization"),
        },
    ),
]

model = FewShotNer()
model.contextualize(entities=entities, examples=examples)
model.predict(["Pedro Pereira is the president of Peru and the owner of Walmart."])

# OUTPUT:

    AnnotatedDocument(
        text="Pedro Pereira is the president of Peru and the owner of Walmart.",
        annotations={
            Annotation(start=0, end=13, label="person"),
            Annotation(start=34, end=39, label="location"),
            Annotation(start=56, end=63, label="organization"),
        },
    ),

\end{python}
\caption{Example usage of \texttt{llmNER} for few-shot NER.}
	\label{fig:example_usage2}
\end{figure}

\section{Illustrative examples}

To demonstrate the library's functionality and usefulness, we assessed its performance on two public datasets in English and Spanish: CoNLL 2003 and CoNLL 2002. We measured the performance in all four entity classes: LOC (Location), MISC (Miscellaneous), ORG (organization), and PER (Person). Currently, the state-of-the-art (SOTA) for both benchmarks is a fine-tuned method proposed by \citet{wang-etal-2021-automated}.

We solved both benchmarks with zero-shot and few-shot paradigms using multiple LLMs with all of the hyperparameter combinations available in our library and compared the performance of each combination. We measured the relaxed $F_1$ score by each entity class in the benchmark.

Tables \ref{tab:conll2003} and \ref{tab:conll2002} present the results of the CoNLL 2003 and CoNLL 2002 NER tasks, utilizing the models and some hyperparameter combinations exposed by the library. 

Both tables show the superior performance of GPT 3.5 across both datasets, as this model is the most tested across literature and achieves state-of-the-art in multiple tasks \cite{ye2023comprehensive}. As expected, incorporating few-shot examples into the prompts led to a general performance enhancement because it is known that LLM's performance scales with the number of examples given in the prompt \cite{Song2023, brown2020}.

Regarding the prompting method we used. For the English dataset, the best-performing method was the multi-turn for zero-shot learning, a result that \citet{xie2023empirical} also reported. The best prompting method for few-shot learning was the single-turn, but only by a small margin. This situation can be attributed to the fact that the most decisive contribution to the model's decision is the few-shot examples, not the prompting method. For the Spanish dataset, the best performant prompting strategy for zero-shot learning was single-turn by a small margin and multi-turn for few-shot learning. This result shows that the performance may vary across datasets, and the user should perform hyperparameter optimization for their experiments.

Considering the answer shape, our proposed in-line answer shape generally outperforms JSON; this is important because almost all the literature shapes the answers as JSON and does not explore multiple answer shapes in the prompt engineering phase.

In general, the models performed better for the English NER benchmark (CoNLL 2003); most of the text on the LLM's training \textit{corpora} are in English; therefore, it is easier for those models to solve tasks in the English language.

Results show that the SOTA performance on the respective benchmarks, represented by the ACE model \cite{wang-etal-2021-automated}, outperforms all our results by a significant margin. Despite our comparative standing being considerably lower than the SOTA, we exploit a different paradigm. The SOTA model exploits the model fine-tuning paradigm, which uses a task-specific training set to adapt the model parameters for the specific downstream task, and we are using a general-knowledge model, only fine-tuned for instruction following to solve a downstream task by only describing the task in natural language; hence both results cannot be easily compared. The main contribution of our present work and our main goal is not the performance of the strategies but the release of a library to ease the research of NER through in-context learning.

\setlength{\tabcolsep}{3pt}
\begin{table*}[]
    \begin{adjustbox}{width=0.7\textwidth,center}
\begin{tabular}{ccclccccc}
\hline
\multicolumn{4}{c}{\textbf{Model}} & \multicolumn{1}{c}{\textbf{LOC}} & \multicolumn{1}{c}{\textbf{MISC}} & \multicolumn{1}{c}{\textbf{ORG}} & \multicolumn{1}{c}{\textbf{PER}} & \multicolumn{1}{c}{\textbf{\textmu}$\mathbf{F_1}$} \\ \hline
\multirow{12}{*}{Zero-shot} & \multirow{6}{*}{Single-turn} & \multirow{3}{*}{JSON} & GPT 3.5 & 67.0 & 29.0 & 51.9 & 79.9 & 62.9 \\
 &  &  & Mixtral 8x7B & 44.0 & 7.3 & 29.0 & 55.2 & 36.8 \\
 &  &  & Llama 2 70B & 38.6 & 4.5 & 37.1 & 58.4 & 33.5 \\ \cline{3-9} 
 &  & \multirow{3}{*}{In-line} & GPT 3.5 & 62.1 & 14.6 & 59.1 & 89.3 & 59.8 \\
 &  &  & Mixtral 8x7B & 52.3 & 4.8 & 34.6 & 73.6 & 39.2 \\
 &  &  & Llama 2 70B & 28.3 & 2.6 & 18.6 & 28.5 & 23.5 \\ \cline{2-9} 
 & \multirow{6}{*}{Multi-turn} & \multirow{3}{*}{JSON} & GPT 3.5 & 74.3 & 3.1 & 41.1 & 90.1 & 68.8 \\
 &  &  & Mixtral 8x7B & 47.0 & 5.8 & 37.1 & 56.4 & 41.0 \\
 &  &  & Llama 2 70B & 41.6 & 4.8 & 27.6 & 77.1 & 46.0 \\ \cline{3-9} 
 &  & \multirow{3}{*}{In-line} & GPT 3.5 & 76.1 & 28.3 & 52.4 & 81.2 & \textbf{72.9} \\
 &  &  & Mixtral 8x7B & 43.6 & 2.7 & 17.8 & 57.2 & 36.0 \\
 &  &  & Llama 2 70B & 35.2 & 1.8 & 16.3 & 41.7 & 31.2 \\ \hline\hline
\multirow{12}{*}{Few-shot} & \multirow{6}{*}{Single-turn} & \multirow{3}{*}{JSON} & GPT 3.5 & 65.7 & 34.2 & 55.6 & 80.6 & 62.6 \\
 &  &  & Mixtral 8x7B & 52.7 & 15.2 & 36.0 & 77.8 & 46.9 \\
 &  &  & Llama 2 70B & 49.6 & 13.7 & 31.5 & 79.6 & 48.3 \\ \cline{3-9} 
 &  & \multirow{3}{*}{In-line} & GPT 3.5 & 75.5 & 48.9 & 62.3 & 88.8 & \textbf{73.5} \\
 &  &  & Mixtral 8x7B & 59.4 & 20.1 & 51.2 & 78.6 & 54.6 \\
 &  &  & Llama 2 70B & 54.5 & 19.0 & 38.8 & 62.8 & 48.1 \\ \cline{2-9} 
 & \multirow{6}{*}{Multi-turn} & \multirow{3}{*}{JSON} & GPT 3.5 & 60.3 & 17.4 & 59.4 & 75.3 & 58.0 \\
 &  &  & Mixtral 8x7B & 47.2 & 15.8 & 31.3 & 79.7 & 54.6 \\
 &  &  & Llama 2 70B & 59.3 & 7.8 & 48.1 & 86.9 & 57.0 \\ \cline{3-9} 
 &  & \multirow{3}{*}{In-line} & GPT 3.5 & 67.8 & 17.6 & 51.3 & 88.2 & 72.1 \\
 &  &  & Mixtral 8x7B & 40.2 & 2.0 & 12.6 & 36.8 & 30.4 \\
 &  &  & Llama 2 70B & 25.8 & 3.5 & 16.9 & 30.6 & 20.9 \\ \hline
\end{tabular}
    \end{adjustbox}
	\caption{\texttt{llmNER} $F_1$ performance results (in percentage) over CoNLL 2003 (English NER task). \textbf{\textmu}$\mathbf{F_1}$ column represents micro-averaged $F_1$ over all entity classes.}
	\label{tab:conll2003}
\end{table*}

\begin{table*}[]
    \small
	\centering
	\begin{adjustbox}{width=0.7\textwidth,center}
\begin{tabular}{ccclccccc}
\hline
\multicolumn{4}{c}{\textbf{Model}} & \multicolumn{1}{c}{\textbf{LOC}} & \multicolumn{1}{c}{\textbf{MISC}} & \multicolumn{1}{c}{\textbf{ORG}} & \multicolumn{1}{c}{\textbf{PER}} & \multicolumn{1}{c}{\textbf{\textmu}$\mathbf{F_1}$} \\ \hline
\multirow{12}{*}{Zero-shot} & \multirow{6}{*}{Single-turn} & \multirow{3}{*}{JSON} & GPT 3.5 & 73.4 & 8.3 & 56.2 & 85.1 & \textbf{64.0} \\
 &  &  & Mixtral 8x7B & 50.9 & 7.8 & 49.2 & 59.2 & 44.9 \\
 &  &  & Llama 2 70B & 56.3 & 6.4 & 38.1 & 74.8 & 39.6 \\ \cline{3-9} 
 &  & \multirow{3}{*}{In-line} & GPT 3.5 & 68.9 & 10.8 & 56.8 & 90.3 & 57.2 \\
 &  &  & Mixtral 8x7B & 47.7 & 4.3 & 36.3 & 68.2 & 35.4 \\
 &  &  & Llama 2 70B & 37.4 & 1.7 & 10.3 & 15.7 & 20.0 \\ \cline{2-9} 
 & \multirow{6}{*}{Multi-turn} & \multirow{3}{*}{JSON} & GPT 3.5 & 67.6 & 6.6 & 59.1 & 75.3 & 55.4 \\
 &  &  & Mixtral 8x7B & 47.4 & 0.0 & 39.5 & 56.7 & 42.0 \\
 &  &  & Llama 2 70B & 52.6 & 6.3 & 54.0 & 64.2 & 39.9 \\ \cline{3-9} 
 &  & \multirow{3}{*}{In-line} & GPT 3.5 & 60.6 & 20.7 & 66.7 & 78.5 & 63.7 \\
 &  &  & Mixtral 8x7B & 50.2 & 4.4 & 52.8 & 58.8 & 39.6 \\
 &  &  & Llama 2 70B & 27.5 & 3.7 & 22.9 & 33.1 & 23.0 \\ \hline\hline
\multirow{12}{*}{Few-shot} & \multirow{6}{*}{Single-turn} & \multirow{3}{*}{JSON} & GPT 3.5 & 64.6 & 19.8 & 52.1 & 85.1 & 59.0 \\
 &  &  & Mixtral 8x7B & 60.3 & 14.8 & 38.6 & 78.3 & 47.3 \\
 &  &  & Llama 2 70B & 66.9 & 12.7 & 23.5 & 77.1 & 41.1 \\ \cline{3-9} 
 &  & \multirow{3}{*}{In-line} & GPT 3.5 & 64.5 & 15.4 & 54.4 & 87.2 & 61.4 \\
 &  &  & Mixtral 8x7B & 75.5 & 15.6 & 63.0 & 80.2 & 56.5 \\
 &  &  & Llama 2 70B & 52.0 & 17.5 & 50.5 & 62.9 & 49.1 \\ \cline{2-9} 
 & \multirow{6}{*}{Multi-turn} & \multirow{3}{*}{JSON} & GPT 3.5 & 71.6 & 18.2 & 61.3 & 88.7 & 58.5 \\
 &  &  & Mixtral 8x7B & 54.7 & 13.1 & 41.7 & 79.8 & 45.1 \\
 &  &  & Llama 2 70B & 54.3 & 9.2 & 33.2 & 60.0 & 34.5 \\ \cline{3-9} 
 &  & \multirow{3}{*}{In-line} & GPT 3.5 & 67.8 & 15.2 & 72.2 & 92.6 & \textbf{68.8} \\
 &  &  & Mixtral 8x7B & 38.9 & 4.0 & 25.3 & 48.0 & 29.3 \\
 &  &  & Llama 2 70B & 43.2  & 7.5 & 0.0 & 53.7 & 25.6  \\ \hline
\end{tabular}
    \end{adjustbox}
	\caption{\texttt{llmNER} $F_1$ performance results (in percentage) over CoNLL 2002 (Spanish NER task). \textbf{\textmu}$\mathbf{F_1}$ column represents micro-averaged $F_1$ over all entity classes.}
	\label{tab:conll2002}
\end{table*}

\section{Impact}
\texttt{llmNER}'s main characteristic is its ease of use for performing zero-shot and few-shot NER alongside its flexible interface for performing prompt engineering. Regarding those strengths, our library can help push the boundaries of ICL research by removing the barrier of the prompting and parsing steps. Without this barrier, ICL can be easily used in applied machine learning, such as LLM research on clinical NLP, improving the research in areas adjacent to machine learning and NLP.

Our library has been used on clinical NLP as a pre-annotation step for streamlining human annotations of corpora for privacy-preserving clinical NLP \cite{aracena2024, Dunstan2024}.

\section{Conclusions}

The developed Python library offers an accessible framework for conducting zero and few-shot NER, validated in Spanish and English using CoNLL 2003 and CoNLL 2002 benchmarks. This comprehensive tool encompasses various prompting methods, answer shape parsers, and POS augmentation techniques, allowing researchers to explore diverse prompt engineering configurations.

Our results are not comparable with the SOTA because our approach diverges significantly from conventional NER methodologies. Rather than relying on task-specific fine-tuning, we leverage a general-knowledge model, fine-tuned solely for instruction following. Consequently, direct comparison with SOTA results is challenging due to the inherent paradigm differences.

The release of the library to the research community aims to foster reproducibility and innovation in NLP research, enabling researchers to explore and validate various settings, thus advancing the collective understanding of NER through in-context learning.

\section*{Acknowledgements}
This work was funded by ANID Chile National Doctoral Scholarships 21211659 (CA) and 21220200 (FV).





\bibliographystyle{elsarticle-num-names} 
\bibliography{bibliography}

\end{document}